\documentclass[10pt,twocolumn,letterpaper]{article}

\usepackage[pagenumbers]{cvpr}   

\usepackage{multirow}
\usepackage{makecell}
\usepackage{array}
\usepackage{colortbl}

\definecolor{mygray}{gray}{0.92}
\newcommand{\thickhline}{\noalign{\hrule height 1.2pt}}

\definecolor{cvprblue}{rgb}{0.21,0.49,0.74}
\usepackage[pagebackref,breaklinks,colorlinks,allcolors=cvprblue,hypertexnames=false]{hyperref}

\def\paperID{*****} 
\def\confName{CVPR}
\def\confYear{2026}

\title{PromptPath: Prompt-Adaptive Computational Pathways for \\In-Context Learning}

\author{
  Hangrui Zhang$^{1}$, \;Feifei Shao$^{1}$\thanks{Feifei Shao is the corresponding author.}\;, Yawei Luo$^1$, \;Ping Liu$^{2}$, \;Jiaxiang Liu$^{3}$, \\
  Zuoqi Tang$^{1}$, \;Zhao Wang$^1$, \;Hongwei Wang$^1$, \;Jun Xiao$^1$ \\
  \small $^1$ Zhejiang University, China\; 
  \small $^2$ University of Nevada, Reno, USA \; \\
  \small $^3$ Guangdong Institute of Intelligence Science and Technology, China \; \\
  \small \texttt{\{hanryz, sff, yaweiluo, tangzq, zhao\_wang\}@zju.edu.cn, pino.pingliu@gmail.com, } \\
  \small \texttt{forworkliu@gmail.com, hongweiwang@intl.zju.edu.cn, junx@cs.zju.edu.cn} \\
  }

\begin{document}
\maketitle

\begin{abstract}
In-context learning (ICL) has attracted increasing attention for enabling models to perform new tasks using only a few ``input--output'' prompt examples. However, existing approaches suffer from \textbf{shallow task adaptation}, where prompts are primarily used as contextual cues to implicitly infer task intent through semantic representations, while the underlying computational process remains unchanged. This limitation restricts task-specific adaptation and compromises inference interpretability. We argue that prompts should not only condition feature representations but also dynamically regulate the model's computation pathways. To this end, we propose \textbf{PromptPath}, an adaptive ICL framework that enables computation-level adaptation through prompt-conditioned dynamic pathways. Specifically, PromptPath introduces a prompt-driven routing mechanism to selectively activate and compose lightweight low-rank experts, forming task-specific computational pathways tailored to different prompts. By integrating prompt information directly into the inference process, PromptPath dynamically reconfigures model computation to enhance task specialization and interpretability. Extensive experiments on 3D point cloud and 2D visual recognition benchmarks demonstrate that PromptPath consistently outperforms state-of-the-art ICL baselines while exhibiting strong cross-domain and cross-task generalization.

\end{abstract}

\section{Introduction}

\begin{figure}[t]
  \centering
  \includegraphics[width=1.0\linewidth]{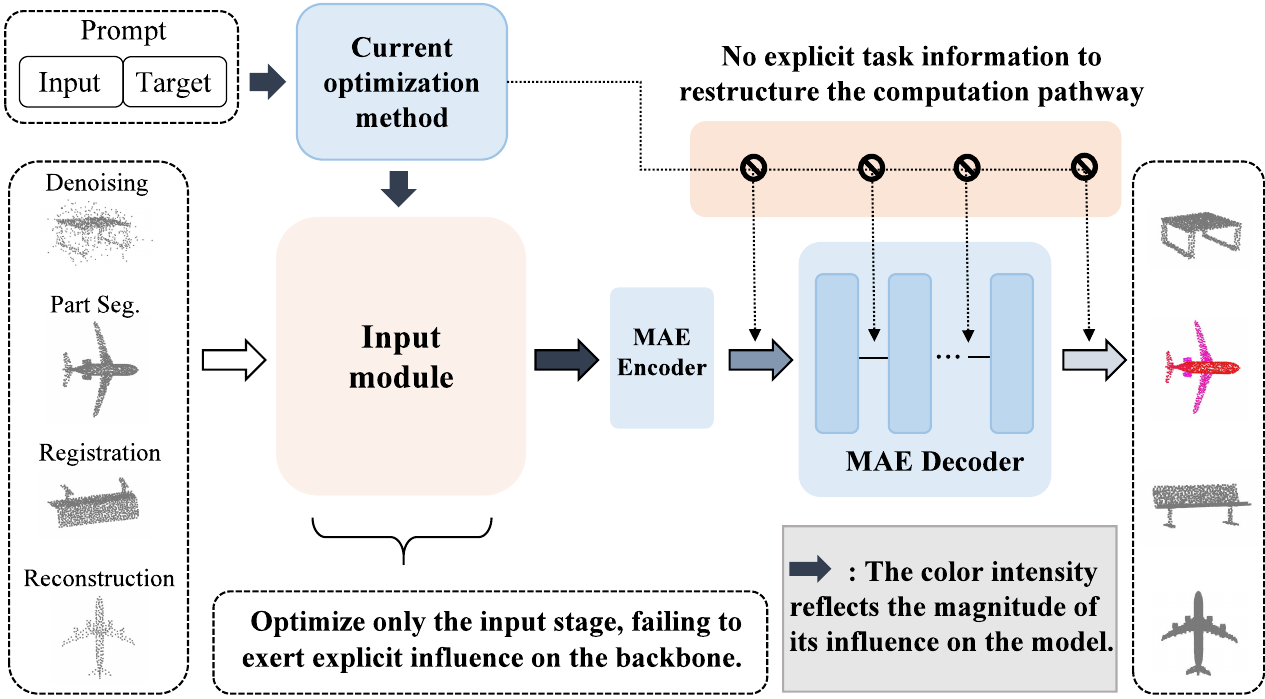}
  \caption{Existing ICL methods mainly optimize input-stage prompts while keeping computational pathway fixed, causing implicit task guidance. Moreover, prompt influence is strong in shallow layers but fades in deeper ones~\cite{rela12}.}
  \label{intro}
\end{figure}

Recently, with the rapid advancement of large-scale models, in-context learning (ICL)~\cite{nlp1,nlp2,nlp3} has emerged as a promising paradigm that enables models to perform new tasks on the fly by conditioning on a few in-context examples---called prompts---rather than updating parameters explicitly. The model learns task mappings from these prompt pairs and directly applies them to query pairs within a single forward pass~\cite{zhou2022learning,zhou2022conditional,yao2023visual}. By leveraging only a few prompts, ICL allows a single model to generalize across diverse tasks without fine-tuning, demonstrating remarkable scalability and adaptability in large-model applications.

Prompts act as task-defining cues that guide model reasoning and have played a central role in ICL advances. Consequently, much research has focused on selecting optimal prompts, either via similarity-based retrieval, which favors prompts closely related to the query~\cite{liu2021makes, rubin2021learning, nie2022cross, li2023unified}, or diversity-based retrieval, which reduces redundancy and ensures broader coverage~\cite{zhang2022automatic, yu2022generate, levy2022diverse}. More recent methods, such as Condenser~\cite{Condenser}, aggregate relevant visual–context pairs into latent representations to improve reasoning precision. However, they concentrate on enhancing prompt representations rather than reshaping the model’s internal computation, constraining the emergence of truly task-adaptive processing behaviors.

We argue that this \textbf{shallow task adaptation issue} stems from the limited integration of task information during inference. Specifically, existing ICL methods mainly treat prompts as contextual inputs that specify the task but do not explicitly modulate the computational process~\cite{he2022mae,painter}. For example, Yin et al.~\cite{yin2025lifting} show that prompts are mainly injected into semantic tokens in early layers, while Petrov et al.~\cite{petrov2024when} demonstrate that prompting and ICL have limited influence on internal attention computation. As a result, prompt-induced task information is primarily captured by early layers and gradually attenuated in deeper layers, which is shown in Figure~\ref{intro}.  Although different prompts may lead to different inference behaviors, such variations are largely reflected at the feature representation level rather than through explicit and structured modifications of the computation pathway. Therefore, the model still relies on largely fixed internal pathways for inference, leading to shallow task adaptation and limited interpretability and controllability.

To address this limitation, we propose a different perspective: prompts should guide not only input representations but also the internal computation itself. This motivates a shift from implicit representation-level conditioning to explicit computation-level adaptation, in which the model’s inference pathway is dynamically reconfigured by the prompt. To realize this, we propose \textbf{PromptPath}, a prompt-adaptive ICL framework that enables dynamic inference pathways conditioned on the task prompt. As shown in Figure~\ref{framework}, PromptPath employs a prompt-driven routing mechanism to activate task-relevant experts from a lightweight expert library and integrate them into the model in a structured manner. This design enables the model to dynamically reconfigure its internal computational pathways, resulting in distinct and interpretable task-specific behaviors while making prompt-driven adaptation explicit rather than implicit.

Finally, to comprehensively evaluate the generalization capability of PromptPath, we conduct experiments across diverse visual domains and tasks, including 3D point cloud understanding and 2D image recognition. We further evaluate its cross-dataset and cross-task generalization in Table~\ref{tab:sota_ni} and Table~\ref{tab:heldout_task}, respectively. Moreover, PromptPath achieves consistent performance improvements, surpassing MICAS by up to 2.7 mIoU points on 3D Part Segmentation and Condenser by 2.03 mIoU points on 2D foreground segmentation.
The contributions of this paper are summarized as follows:
\begin{itemize}
\item To the best of our knowledge, we are among the first to reveal that existing ICL relies on implicit, representation-level prompt conditioning and argue that prompts should guide both representation learning and the model’s internal computational processes.
\item We propose PromptPath, a prompt-adaptive ICL framework that introduces computation-level adaptation by dynamically constructing task-specific inference pathways via prompt-driven routing, enabling explicit, interpretable, and flexible task specialization.
\item Extensive experiments on diverse 3D point cloud and 2D image recognition benchmarks demonstrate that PromptPath consistently outperforms state-of-the-art ICL methods while maintaining strong generalization.
\end{itemize}

\section{Related Work}

\subsection{Representation-Level Task Conditioning}

Visual in-context learning (ICL) adapts pretrained models from a few example pairs without parameter updates~\cite{nlp1,nlp2,nlp3}. Existing frameworks concatenate support examples with the query~\cite{zhou2022learning,zhou2022conditional,yao2023visual}, improve prompt selection through similarity-based~\cite{liu2021makes,rubin2021learning,nie2022cross,li2023unified} or diversity-based retrieval~\cite{zhang2022automatic,yu2022generate,levy2022diverse}, or construct latent prompts~\cite{Condenser}. Unified frameworks such as Taskonomy~\cite{Taskonomy}, Unified-IO~\cite{Lu_2024_CVPR}, Painter~\cite{painter}, and Kosmos-2~\cite{kosmos2} cover heterogeneous visual tasks, while PointPrompt~\cite{pointprompt} and 3D-Adapter~\cite{chen2025dadapter} extend prompt-based adaptation to point clouds.

These methods inject task cues at the \textit{representation level}, through input embeddings or early activations~\cite{rela11,zhang2024instruct}, while retaining a fixed computation pathway. Consequently, adaptation remains an implicit activation shift, limiting task-specific processing and interpretability.

\subsection{MoE for Dynamic Inference Pathways}

Mixture-of-Experts (MoE) enables dynamic inference by routing inputs to specialized modules. VisionMoE~\cite{riquelme2021scaling} and Switch Transformer~\cite{fedus2021switch} demonstrate efficient, scalable conditional computation; subsequent work improves routing stability and generalization~\cite{soft_moe_2023}, and Point-MoE~\cite{chen2025point} extends expert specialization to 3D vision. Mixture-of-LoRA and LoRA-MoE similarly gate low-rank modules for parameter-efficient adaptation and expert fusion~\cite{dou2024loramoe,feng2024mixture}.

However, most MoE methods route intermediate content features~\cite{rela21} rather than explicit task semantics~\cite{rela22}, learning expert specialization during training instead of inferring it from prompts at inference time~\cite{rela23}. PromptPath addresses this limitation by exploiting transformations encoded in visual prompt pairs to compose low-rank experts and dynamically reconfigure computation pathways for prompt-defined tasks.

\section{Methodology}

\begin{figure*}[t]
  \centering
  \includegraphics[width=1.0\linewidth]{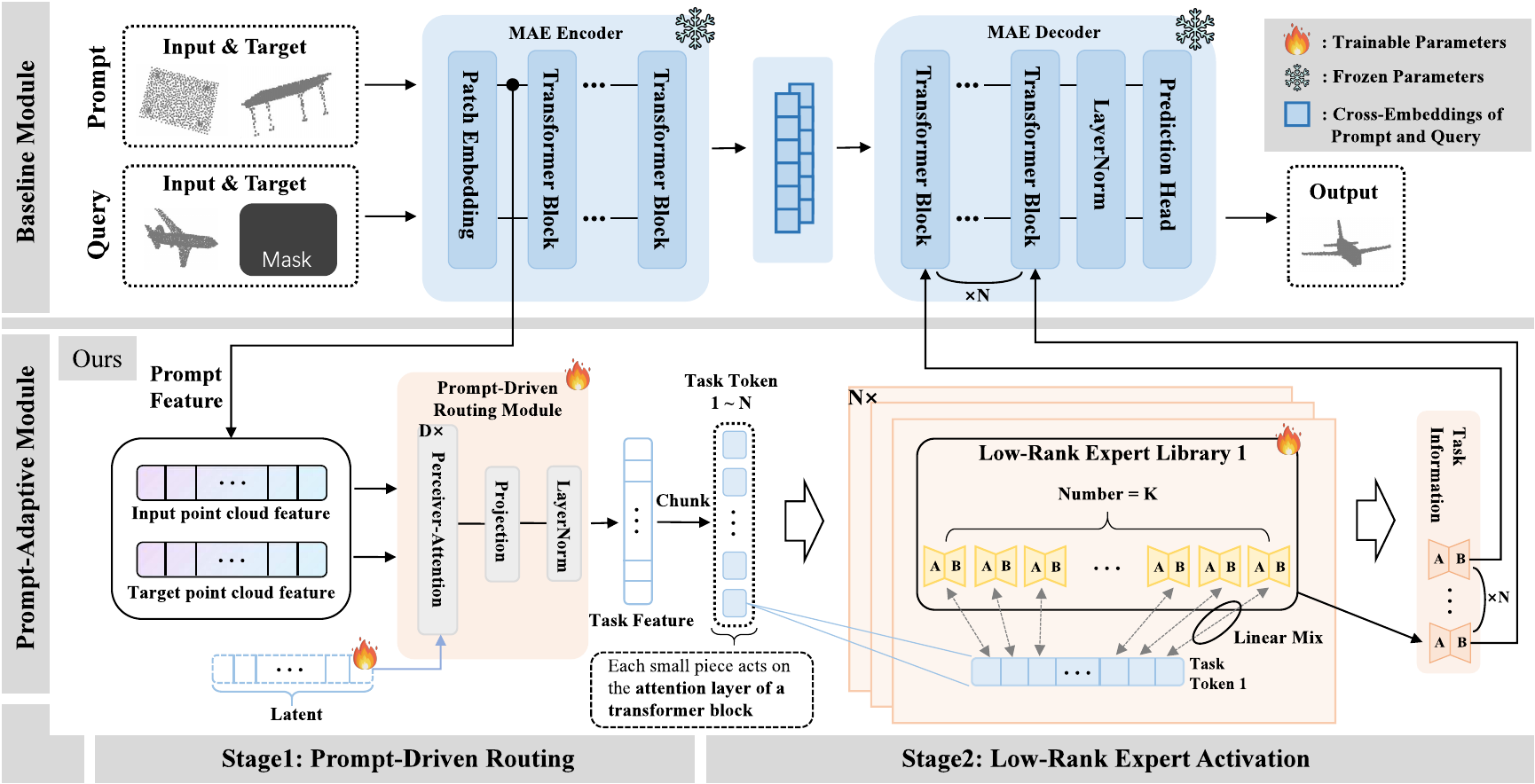}
  \caption{Overview of the PromptPath framework, which consists of a prompt-driven routing (\textbf{PDR}) mechanism and low-rank expert library (\textbf{LoRE}). The former computes activation signals for experts based on the given prompt to foster genuine task understanding, while the latter activates the experts and provides task-specific guidance throughout the entire inference process.}
  \label{framework}
\end{figure*}

\subsection{Problem Setting}

We consider the in-context learning (ICL) setting, where a model is provided with one or more input–target example pairs as prompts, and must infer the correct target for a new query \textit{without modifying its parameters}. 
A prompt pair is denoted as $P = (X_p, Y_p)$ and a query as $Q = (X_q, Y_q)$. 
During inference, the model predicts the query target by conditioning on the prompt:
\begin{equation}
\hat{Y}_q = f_{\theta}(X_q \mid X_p, Y_p),
\label{eq:icl_inference}
\end{equation}
with model parameters $\theta$ fixed.

Following standard practice,  we assume the model has been preconditioned using a masked reconstruction objective in the style of Masked Autoencoders (MAE)~\cite{he2022masked}, allowing it to internalize both spatial semantics within instances and transformation semantics between input–target pairs. 
Under this setup, task adaptation at inference occurs through conditioning on the prompt as in Eq.~\ref{eq:icl_inference}; the pretrained backbone parameters remain fixed.
However, this adaptation operates only at the representation level: the prompt influences embeddings or early feature activations, while the internal computation pathway defined by $\theta$ remains fixed. 
This observation motivates us to enable prompts to modulate not only feature embeddings, but also the \emph{computation pathways} executed during inference.


\subsection{PromptPath Overview}

Our goal is to let the prompt influence not only input representations but also the \emph{internal computation pathways} executed during inference. 
In standard ICL, prompts essentially modulate embeddings while the computational structure defined by $\theta$ remains fixed, so prompts merely affect \textit{what} features are emphasized rather than \textit{how} the query is processed.

PromptPath introduces {computation-level adaptation}, where the prompt configures the inference pathway itself:
\begin{equation}
\hat{Y}_q = f_{\theta,\,E(P)}(X_q \mid X_p, Y_p),
\label{eq:2}
\end{equation}
where $\theta$ remains fixed and $E(P)$ is a prompt-adaptive modulation assembled from a shared expert library. 
This makes the computation pathway prompt-dependent, enabling explicit, task-aware behaviors during inference while keeping the pretrained backbone fixed.
As shown in Figure~\ref{framework}, PromptPath consists of two key components: prompt-driven routing mechanism and low-rank expert library.

For clarity of exposition, we describe these components in the context of 3D point cloud ICL, where input–target pairs naturally follow the formulation in Eq.~\ref{eq:2}. 
However, the procedure is modality-agnostic and applies equally to 2D images and other domains, since PromptPath modulates the computation pathway rather than relying on modality-dependent architectural assumptions.


\subsubsection{Prompt-Driven Routing Mechanism}

The prompt-driven routing (PDR) mechanism extracts high-level task semantics from the prompt pair \(P = (X_p, Y_p)\) and translates them into routing signals for dynamic expert activation, enabling task-aware reconfiguration of the model's computation pathway.

We adopt a Perceiver-style attention~\cite{jaegle2021perceiver} to model the relational structure within the prompt. 
The input and target are encoded by the backbone into embeddings \(E_{X_p}\) and \(E_{Y_p}\), which are projected into a latent space and processed by a stack of \(D\) Perceiver-Attention $\Phi_{PA}$ blocks:
\begin{equation}
F_{prompt} = \Phi_{PA}^{D} \circ \cdots \circ \Phi_{PA}^{1}(E_{X_p}, E_{Y_p}),
\end{equation}
where “$\circ$” denotes iteratively that one block’s output becomes the next block’s input. 
Each block performs latent self-attention followed by cross-attention with \(E_{X_p}\) and \(E_{Y_p}\), capturing both structural and transformation semantics.

The output \(F_{prompt}\) is then projected and normalized to form a compact task feature \(F_{task}\), which is subsequently split into a sequence of task tokens \(T_{task}\). 
These tokens parameterize a routing function that dynamically activates and composes experts from the low-rank expert library to form a task-specific computation path. We inject this modulation into the decoder attention projections, whose Q/K/V/O transformations directly govern prompt--query interactions during task-specific output generation. Keeping the encoder unchanged preserves its pretrained visual representations while localizing adaptation to the decoder stage.
In this way, the prompt determines the internal inference pathway without altering the underlying model parameters.

\subsubsection{Low-Rank Expert Library}

The low-rank expert library (LoRE) provides a set of modular, task-specialized computations that are dynamically integrated into the backbone during inference. 
Since the MAE decoder contains \(N\) Transformer blocks, we instantiate \(N\) corresponding expert libraries and split the task feature \(F_{task}\) into \(N\) task tokens \(\{T_{task}^n\}_{n=1}^{N}\), where each \(T_{task}^n\) serves as a routing vector \(\alpha^n \in \mathbb{R}^{K}\) for the \(n\)-th library containing \(K\) experts.

Each expert \(\varepsilon_i^n\) in the \(n\)-th library adopts the Low-Rank Adaptation (LoRA)~\cite{hu2022lora} form:
\begin{equation}
\begin{aligned}
\varepsilon_i^n &= B_i^n A_i^n, \quad A_i^n \in \mathbb{R}^{r \times d_2}, \; B_i^n \in \mathbb{R}^{d_1 \times r}, \\
&\quad r \ll \min(d_1, d_2).
\end{aligned}
\end{equation}
Vector \(\alpha^n\) composes experts into a task-dependent module:
\begin{equation}
\bar{\varepsilon}^n = \bar{B}^n \bar{A}^n,
\end{equation}
where
\begin{equation}
\bar{A}^n = \sum\nolimits_{i=1}^{K} \alpha_i^n A_i^n, \quad
\bar{B}^n = \sum\nolimits_{i=1}^{K} \alpha_i^n B_i^n.
\end{equation}
Rather than altering the block's parameters, the composed expert $\bar{\varepsilon}^n$ is \textit{externally injected during inference} to modulate the block's computation path for the given prompt.
Given the original weight matrix \(W_0 \in \mathbb{R}^{d_1 \times d_2}\) and input \(F_{in}\), the modified output $F_{out}$ is:
\begin{equation}
F_{out} = W_0 F_{in} + \bar{B}^n \bar{A}^n F_{in}
           = (W_0 + \Delta W^n)F_{in},
\end{equation}
where \(\Delta W^n = \bar{B}^n \bar{A}^n\).

Notably, the backbone parameters \(W_0\) remain fixed. 
The composed expert \(\Delta W^n\) adjusts the computation pathway in a prompt-adaptive manner, realizing the modulation function \(E(P)\) (cf. Eq.~\ref{eq:2}) and allowing the model to dynamically adapt inference behavior based on task semantics without parameter fine-tuning.

\subsection{Learning Prompt-Adaptive Modulation}

PromptPath is optimized in a lightweight manner: the pretrained backbone remains completely {frozen}, and only the prompt-driven routing module and the low-rank expert library are trained. 
This ensures that flexible {computation-level adaptation} is achieved through prompt-adaptive modulation of the inference pathway, rather than parameter fine-tuning.

\subsubsection{3D Point Cloud ICL Setting.}
We adopt MICAS~\cite{MICAS} as the baseline model. 
Following its standard setup, the optimization objective is computed as:
\begin{equation}
\mathcal{L}_{\text{micas}} =
\mathcal{L}_{\text{CD}}(f_{\theta,\,E(P)}(X_q \mid X_p, Y_p), \ Y_q),
\end{equation}
where $\mathcal{L}_{\text{CD}}$ is the Chamfer Distance~\cite{fan2017point}, which measures the geometric discrepancy between the predicted point cloud $\hat{Y}_q$ (cf. Eq.~\ref{eq:2}) and its corresponding ``target'' point cloud $Y_q$. 



\subsubsection{2D Image ICL Setting.}
For 2D image-based in-context learning, we adopt Condenser~\cite{Condenser}. 
The overall optimization objective combines a token prediction loss $\mathcal{L}_{\text{TP}}$ and a feature pre-alignment loss $\mathcal{L}_{\text{PA}}$ as follows:
\begin{equation}
\mathcal{L}_{\text{condenser}} =  
\mathcal{L}_{\text{TP}} + \lambda \cdot \mathcal{L}_{\text{PA}},
\end{equation}
where
\begin{equation}
\begin{aligned}
\mathcal{L}_{\text{TP}} & =  
\mathcal{L}_{\text{CE}}(g_{\theta,\,E(P)}(X_q \mid X_p, Y_p), \ \dot{Y}_q),\\
\mathcal{L}_{\text{PA}} & =  
\mathcal{L}_{\text{CS}}(h_{\theta,\,E(P)}(X_p, Y_p), \ (\ddot{X}_q, \ddot{Y}_q)),
\end{aligned}
\end{equation}
where $\mathcal{L}_{\text{CE}}$ and $\mathcal{L}_{\text{CS}}$ denote the cross-entropy and cosine similarity losses, respectively. The function $g(\cdot)$ represents the token generator, which outputs image tokens, while $h(\cdot)$ denotes the feature generator that produces image features. $\dot{(\cdot)}$ and $\ddot{(\cdot)}$ denote the tokenization by the VQGAN~\cite{esser2021taming} encoder and the feature extraction by CLIP~\cite{radford2021learning}, respectively. 
The hyperparameter $\lambda$ controls the trade-off between token-level supervision and feature-level alignment.

In summary, PromptPath performs task adaptation through the prompt-adaptive modulation $E(P)$ during inference, while the backbone parameters $\theta$ remain fixed. 
This achieves computation-level adaptation, allowing the model to adjust its processing behavior based on the prompt across tasks.

\section{Experiments}
\begin{table*}[t]
  \centering
  \fontsize{9pt}{11pt}\selectfont
  \setlength\tabcolsep{0.86mm}
  \renewcommand\arraystretch{1.0}
   \caption{Comparison with state-of-the-art models on the ShapeNet In-Context~\cite{fang2024explore}. For reconstruction, denoising, and registration, we report Chamfer Distance (CD)~\cite{fan2017point} loss (x1000). For part segmentation, we report mIoU. PromptPath-*: built on MICAS-*. Boldface indicates the best performance.}
  \begin{tabular}{|r||c|cccccc|cccccc|cccccc|c|}
  \hline\thickhline
  \rowcolor{mygray}
  &  & \multicolumn{6}{c|}{Reconstruction CD $\downarrow$} & \multicolumn{6}{c|}{Denoising CD $\downarrow$} & \multicolumn{6}{c|}{Registration CD $\downarrow$} & Part Seg.  \\
  \rowcolor{mygray}
  \multirow{-2}[-1]{*}{Models}    & \multirow{-2}[-1]{*}{Venues}                  
  & L1    & L2    & L3    & L4    & L5    & \multicolumn{1}{c|}{Avg.} 
  & L1    & L2    & L3    & L4    & L5    & \multicolumn{1}{c|}{Avg.}  
  & L1    & L2    & L3    & L4    & L5    & \multicolumn{1}{c|}{Avg.}  
  & mIoU$\uparrow$              \\ \hline\hline
  \multicolumn{21}{|c|}{Input and target point cloud features are concatenated before entering the backbone}                  \\ \hline
  
  PIC-Cat~\cite{fang2024explore}   
  & NeurIPS'23    
  & \textbf{3.2}   & 3.6   & 4.6   & 4.9   & 5.5   & \textbf{4.3} 
  & 3.9  & 4.6   & 5.3   & 6.0   & 6.8   & 5.3  
  & 10.0   & 11.4   & 13.8  & 16.9  & 18.6 & 14.1 
  & 79.0     \\

  MICAS-Cat~\cite{MICAS}     & CVPR'25  
  & 4.6 & 4.2 & 4.5 & 4.8 & 5.7 & 4.7 
  & 4.2 & 4.4 & 4.6 & 4.9 & 5.1 & 4.6 
  & 5.7 & 6.5 & 9.1 & 12.5 & 15.4 & 9.8 
  & 87.9 \\ 

  PIC++-Cat~\cite{liu2024point}    & IJCV'26   
  & 4.5   & \textbf{3.3}   & \textbf{3.7}   & \textbf{4.6}   & 5.4   & \textbf{4.3}  
  & 3.8  & 4.3   & 5.1   & 5.6   & 6.7   & 5.1   
  & 9.7   & 11.6   & 12.8  & 15.9  & 18.0 & 13.6  
  & 85.3     \\ 

  \textbf{\textit{PromptPath-Cat}} & \textbf{\textit{Ours}}  
  &  3.4 & 4.6 & 4.5 & \textbf{4.6} & \textbf{5.1} & \textbf{4.3} & \textbf{3.7} & \textbf{3.9} & \textbf{4.1} & \textbf{4.4} & \textbf{4.6} & \textbf{4.1}
  & \textbf{4.3} & \textbf{5.1} & \textbf{6.8} & \textbf{9.6} & \textbf{13.3} & \textbf{7.8}
  & \textbf{89.9}  
    \\ \hline
    
    \hline
  \multicolumn{21}{|c|}{Input and target point cloud features are processed independently before entering the backbone}  \\ \hline
  PIC-Sep~\cite{fang2024explore}    & 
  NeurIPS'23    
  & 4.7   & 4.3   & 4.3   & 4.4   & 5.7   & 4.7  
  & 6.3  & 7.2   & 7.9   & 8.2   & 8.6   & 7.6   
  & 8.6   & 9.2   & 10.2  & 11.3  & 12.4 & 10.3 
  & 75.0     \\

  MICAS-Sep~\cite{MICAS}     & CVPR'25 
  & 3.8 & 3.9 & 4.0 & 4.4 & 5.6 & 4.3 
  & 4.4 & 4.9 & 5.2 & 5.5 & 5.7 & 5.1
  & 3.4 & 3.6 & 3.7 & 3.8 & 4.0 & 3.7 
  &  86.8 \\ 

  PIC++-Sep~\cite{liu2024point}   & IJCV'26    
  & 4.5   & 4.1   &  4.2   & 4.2   & 5.9   & 4.6   
  & 5.8  & 7.1   &  7.7   & 8.2   &  8.4   & 7.4   
  & 7.6   &  6.2   & 6.8  & 6.9  & 7.9 &  7.1  
  & 85.5     \\ 
  
  \textbf{\textit{PromptPath-Sep}} & \textbf{\textit{Ours}} 
  & \textbf{2.6} & \textbf{2.8} & \textbf{3.2} & \textbf{3.6} & \textbf{4.6} & \textbf{3.4} 
  & \textbf{2.9} & \textbf{3.2} & \textbf{3.4} & \textbf{3.6} & \textbf{3.7} & \textbf{3.3} 
  & \textbf{2.5} & \textbf{2.7} & \textbf{2.7} & \textbf{2.8} & \textbf{2.9} & \textbf{2.7} 
  &  \textbf{89.5} \\ 

  \hline
  \end{tabular}
  \label{tab:sota_pc}
\end{table*}

\begin{table*}[t]
  \centering
  \fontsize{9pt}{11pt}\selectfont
  \setlength\tabcolsep{0.9mm}
  \renewcommand\arraystretch{1.00}
  \caption{Performance comparison with state-of-the-art models on foreground segmentation, single-object detection, and image colorization (Coloring). The cross-dataset shows that models are trained on coco-5i~\cite{lin2014microsoft} and tested on pascal-5i~\cite{shaban2017one}. $\mathcal{K}$ = 1 and 16 denote the number of used prompts. PromptPath is built on $\text{Condenser}_\text{K=16}$.}
  \begin{tabular}{|>{\raggedleft\arraybackslash}p{49mm}||>{\centering\arraybackslash}p{17.1mm}|ccccc|c|c|c|}
  \hline\thickhline
  \rowcolor{mygray}
   &  & \multicolumn{5}{c|}{Foreground Segmentation mIoU $\uparrow$}
   & Object Detection & Coloring & Cross-dataset \\
  \rowcolor{mygray}
  \multirow{-2}[-1]{*}{Models} & \multirow{-2}[-1]{*}{Venues}
  & L1 & L2 & L3 & L4 & \multicolumn{1}{c|}{Avg.}
  & mIoU $\uparrow$ & MSE $\downarrow$ & mIoU $\uparrow$ \\
  \hline\hline

  Random~\cite{he2022mae}
  & NeurIPS'22 & 28.66 & 30.21 & 27.81 & 23.55 & \multicolumn{1}{c|}{27.56} & 25.45 & 0.67 & - \\
  SupPR~\cite{zhang2023what}
  & NeurIPS'23 & 37.08 & 38.43 & 34.40 & 32.32 & \multicolumn{1}{c|}{35.56} & 28.22 & 0.63 & - \\
  Prompt-SelF~\cite{Voting11}
  & TIP'25 & 35.69 & 38.25 & 35.86 & 33.37 & \multicolumn{1}{c|}{35.79} & 28.08 & 0.63 & 39.66 \\
  Partial2Global~\cite{xu2024towards}
  & NeurIPS'24  & 38.81 & 41.54 & 37.25 & 36.01 & \multicolumn{1}{c|}{38.40} & 30.66 & 0.58 & - \\

  \hline
   $\text{Prompt-SelF}_{\text{w/voting}}$~\cite{Voting11}
  & TIP'25 & 42.48 & 43.34 & 39.76 & 38.50 & \multicolumn{1}{c|}{41.02} & 29.83 & - & -\\
  $\text{Partial2Global}_{\text{w/voting}}$~\cite{xu2024towards} 
  & NeurIPS'24 & 43.23 & 45.50 & 41.79 & 40.22 & \multicolumn{1}{c|}{42.69} & 32.52 & - & - \\

  \hline
  InMeMo~\cite{zhang2024instruct} 
  & WACV'24 & 41.65 & 47.68 & 42.43 & 40.80 & \multicolumn{1}{c|}{43.14} & 43.21 & - & 40.03 \\

  \hline
  $\text{Condenser}_{\mathcal{K}=1}$~\cite{Condenser} 
  & CVPR'25 & 42.13 & 50.31 & 42.20 & 41.90 & \multicolumn{1}{c|}{44.14} & 43.22 & 0.56 & 40.37 \\
  $\text{Condenser}_{\mathcal{K}=16}$~\cite{Condenser} 
  & CVPR'25 & 45.53 & 52.06 & 44.33 & 44.58 & \multicolumn{1}{c|}{46.63} & 44.64 & 0.54 &  40.52 \\

  \hline
  \textbf{\textit{PromptPath}}
    & \textbf{\textit{Ours}}
    & \textbf{46.98} 
    & \textbf{53.53} 
    & \textbf{45.80} 
    & \textbf{48.33} 
    & \multicolumn{1}{c|}{\textbf{48.66}} 
    & \textbf{46.51} 
    & \textbf{0.51} & \textbf{41.41} \\

  \hline
  \end{tabular}
  \label{tab:sota_ni}
\end{table*}

\begin{figure*}[t]
  \centering
  \includegraphics[width=\textwidth]{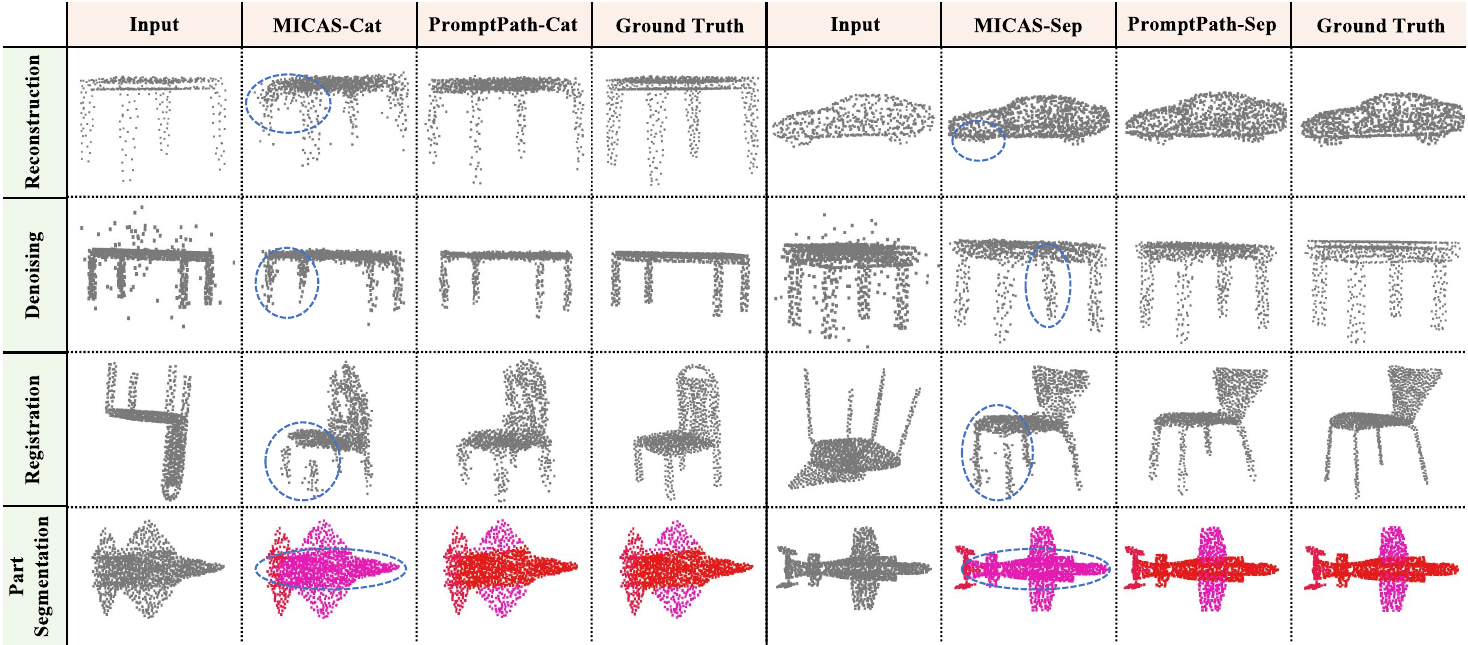}
  \caption{Qualitative results compared with the baseline MICAS. The blue ovals show the significant performance difference observed between PromptPath and baseline MICAS.}
  \label{point}
\end{figure*}

\begin{figure*}[t]
  \centering
  \includegraphics[width=\textwidth]{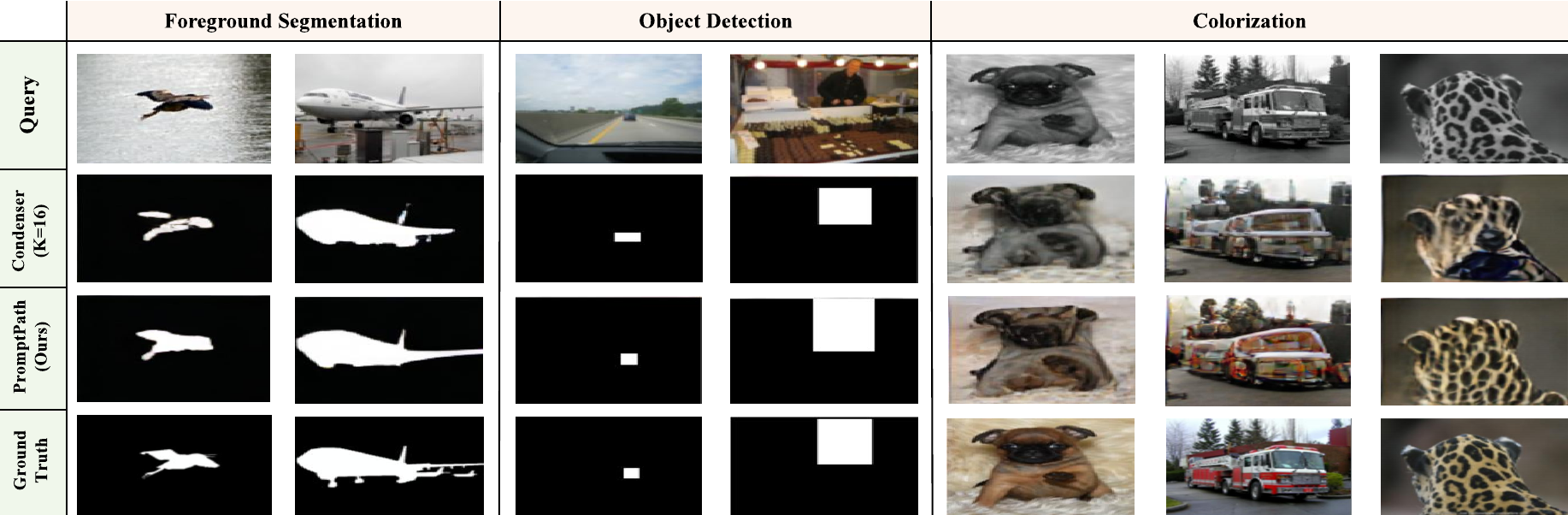}
  \caption{Qualitative results showing performance improvements when compared with baseline Condenser (K=16).}
  \label{image}
\end{figure*}

\subsection{Detailed Experimental Settings}

\label{sec:supp_exp_settings}

PromptPath is evaluated under the experimental settings of MICAS~\cite{MICAS} and Condenser~\cite{Condenser}, which provide representative in-context learning benchmarks for 3D point clouds and 2D images, respectively. Both settings use paired input--target examples and are therefore consistent with the prompt-conditioned formulation used by PromptPath.

\subsubsection{Datasets}

\paragraph{3D point clouds.}
We use the ShapeNet In-Context Dataset introduced by PIC~\cite{fang2024explore}, which is constructed from ShapeNet~\cite{chang2015shapenet} and ShapeNetPart~\cite{yi2016scalable}. It contains 174{,}404 training and 43{,}050 testing point-cloud pairs spanning reconstruction, denoising, registration, and part segmentation. Each task is provided at five difficulty levels.

\paragraph{2D images.}
Following prior work~\cite{he2022mae,xu2024towards,zhang2023what}, we use three standard benchmarks. For few-shot foreground segmentation, Pascal-5\textit{i}~\cite{shaban2017one} contains 20 categories split equally into four folds; the four folds use 2{,}286, 3{,}425, 5{,}883, and 2{,}086 in-context training samples, respectively. For single-object detection, Pascal VOC 2012~\cite{everingham2015pascal} contains 20 categories, and we use 612 in-context training samples. For colorization, we randomly sample 50{,}000 images from the ImageNet-1K~\cite{russakovsky2015imagenet} training set and evaluate on the official validation set. Grayscale images are used as inputs and their color counterparts as targets. For cross-dataset segmentation, models are trained on COCO-5\textit{i}~\cite{lin2014microsoft} and evaluated on Pascal-5\textit{i}.

\subsubsection{Baseline Models}

For 3D point clouds, we build PromptPath on MICAS-Cat and MICAS-Sep~\cite{MICAS}. Both variants follow a masked-autoencoding paradigm and support reconstruction, denoising, registration, and part segmentation. MICAS-Cat concatenates the input and target point-cloud features before the backbone, whereas MICAS-Sep processes them independently before backbone encoding.

For 2D images, we build PromptPath on Condenser~\cite{Condenser}, whose MAE--VQGAN backbone~\cite{he2022mae,esser2021taming} supports foreground segmentation, single-object detection, and colorization. We preserve the original prompt--query conditioning used by each baseline.

\subsubsection{Evaluation Metrics}

We follow the official evaluation protocols of MICAS~\cite{MICAS} and Condenser~\cite{Condenser}. Chamfer Distance (CD)~\cite{fan2017point} measures geometric fidelity for 3D reconstruction, denoising, and registration, while mean Intersection-over-Union (mIoU) evaluates 3D part segmentation. For 2D tasks, mIoU evaluates foreground segmentation and object detection, and Mean Squared Error (MSE) measures colorization accuracy.

\subsubsection{Model Configuration}

All pretrained backbone parameters remain frozen during training; only the prompt-driven routing module and low-rank expert library are optimized. Low-rank task information is injected into the primary attention layer of each decoder Transformer block. Within each attention layer, PromptPath adapts the query, key, value, and output projections, while leaving all other layers unchanged.

The expert library contains $K=1024$ lightweight experts for both MICAS and Condenser. The LoRA rank is 16 for MICAS-Sep, 32 for MICAS-Cat, and 4 for Condenser. In the prompt-driven routing module, the head dimension is set to 64 for MICAS and 1024 for Condenser.

\subsubsection{Training Details}

\paragraph{3D point clouds.}
MICAS-Sep is trained for 150 epochs on three NVIDIA RTX 2080 GPUs with a total batch size of 24. The learning rate follows cosine annealing with a 30-epoch warmup: it starts at $1\times10^{-6}$, reaches $1\times10^{-5}$, and decays to $1\times10^{-6}$. MICAS-Cat is trained on one NVIDIA A100 GPU with batch size 32 and the same schedule, except that its peak learning rate is $4\times10^{-5}$.

\paragraph{2D images.}
All 2D experiments are trained on one NVIDIA A100 GPU. Foreground segmentation, object detection, and colorization use a batch size of 16 and a cyclic cosine learning-rate schedule with a 10-epoch period, oscillating between 0 and 0.03. Cross-dataset foreground segmentation uses a batch size of 64 and the same schedule with a maximum learning rate of 0.06.

\subsection{Comparisons with State-of-the-Art Methods}



\textbf{Results on 3D Tasks.}
We evaluate our approach on the ShapeNet In-Context Dataset~\cite{fang2024explore}, which includes registration, reconstruction, denoising, and part segmentation tasks. Comparisons are made against the state-of-the-art method MICAS~\cite{MICAS}, which has two variants, MICAS-Sep and MICAS-Cat. As shown in Table~\ref{tab:sota_pc} and Figure~\ref{point}, PromptPath consistently and significantly outperforms both MICAS variants across all tasks. Specifically, compared to MICAS-Cat, PromptPath-Cat reduces the registration Chamfer Distance (CD) loss and increases segmentation mIoU by approximately 2 points, while also further lowering the CD loss by 0.5 in the denoising task. Compared to MICAS-Sep, PromptPath-Sep consistently improves performance across all tasks, reducing the CD losses of denoising, reconstruction, and registration by 1.8, 0.9, and 1.0, respectively, and improving segmentation mIoU by 2.7 points. These results clearly demonstrate that composing experts creates dynamic inference pathways that enable more flexible adaptation beyond prior ICL methods.


\textbf{Results on 2D Tasks.}
We evaluate PromptPath under the setting of Condenser~\cite{Condenser}, covering foreground segmentation, single-object detection, and colorization. 
We compare against the full set of baseline categories: single prompt selection~\cite{zhang2023what, xu2024towards}, voting-based inference~\cite{Voting11}, and PEFT-enhanced methods~\cite{zhang2024instruct}, using official results for consistency. 
As shown in Table~\ref{tab:sota_ni} and Figure~\ref{image}, PromptPath achieves the best performance across all tasks. 
It improves segment/detect mIoU by 2.03 and 1.87 compared with Condenser, and reduces MSE for colorization by 0.03, producing more coherent spatial and appearance consistency. 
These results indicate that converting condensed prompts into routing-level expert activation yields more precise and interpretable adaptation than prior ICL methods.

\subsection{Ablation Study}

\begin{table}[t]
  \centering
  \fontsize{9pt}{11pt}\selectfont
  \setlength{\tabcolsep}{4.1pt}
  \renewcommand{\arraystretch}{0.85}
  \caption{Ablation study of our proposed Low-Rank Expert Library (LoRE) and Prompt-Driven Routing (PDR).}
  \begin{tabular}{|c||c|c|c|c|c|c|}
  \hline\thickhline
  \rowcolor{mygray}
  Model & PDR & LoRE &
  \makecell{Rec.\\CD $\downarrow$} &
  \makecell{Den.\\CD $\downarrow$} &
  \makecell{Reg.\\CD $\downarrow$} &
  \makecell{Part Seg.\\mIoU $\uparrow$} \\
  \hline\hline
  \multirow{3}{*}{MICAS-Cat}
  & & & 4.7 & 4.6 & 9.8 & 87.9 \\
  & & $\surd$ & 4.5 & 4.5 & 16.9 & 87.1 \\
  & $\surd$ & $\surd$ &
  \textbf{4.3} &
  \textbf{4.1} &
  \textbf{7.8} &
  \textbf{89.9} \\
  \hline
  \multirow{3}{*}{MICAS-Sep}
  & & & 4.3 & 5.1 & 3.7 & 86.8 \\
  & & $\surd$ & 4.0 & 4.3 & 3.4 & 88.1 \\
  & $\surd$ & $\surd$ &
  \textbf{3.4} &
  \textbf{3.3} &
  \textbf{2.7} &
  \textbf{89.5} \\
  \hline
  \end{tabular}
  \label{tab:ablation}
\end{table}

\begin{table}[t]
  \centering
  \fontsize{9pt}{11pt}\selectfont
  \setlength{\tabcolsep}{5.1pt}
  \renewcommand{\arraystretch}{0.85}
  \caption{Analysis of the expert rank in the Low-Rank Expert Library (LoRE).}
  \begin{tabular}{|c||c|c|c|c|c|}
  \hline\thickhline
  \rowcolor{mygray}
   &  & Rec. & Den. & Reg. & Part Seg. \\
  \rowcolor{mygray}
   \multirow{-2}[-1]{*}{\makebox[17mm][c]{Model}}& \multirow{-2}[-1]{*}{Rank} & CD $\downarrow$ & CD $\downarrow$ & CD $\downarrow$ & mIoU$\uparrow$ \\
  \hline\hline

  \multirow{3}{*}{PromptPath-Cat}
  & 8  & 4.6 & 4.6 & 9.1 & 88.8 \\
  & 16 & 4.4 & 4.3 & 8.7 & 89.3 \\
  & 32 & \textbf{4.3} & \textbf{4.1} & \textbf{7.8} & \textbf{89.9} \\
  
  \hline

  \multirow{3}{*}{PromptPath-Sep}
  & 8  & 3.7 & 4.1 & 3.3 & 88.8 \\
  & 16 & \textbf{3.4} & \textbf{3.3} & \textbf{2.7} & \textbf{89.5} \\
  & 32 & 3.6 & 4.1 & 3.2 & 89.4 \\

  \hline
  \end{tabular}
  \label{tab:analysis_rank}
\end{table}

\textbf{Low-Rank Expert Library (LoRE).} As reported in Table~\ref{tab:ablation}, we augment MICAS-Cat and MICAS-Sep~\cite{MICAS} with LoRE by uniformly averaging all experts without prompt-conditioned routing. For MICAS-Cat, this strategy improves Reconstruction and Denoising but degrades Registration and Part Segmentation. We attribute this behavior to the ``-Cat'' design, which concatenates input and target point cloud representations before feeding them into the model, leading to entangled features that make uniform expert aggregation less effective without prompt-driven routing. In contrast, LoRE consistently benefits MICAS-Sep, reducing Denoising CD by 0.8 and improving Part Segmentation by 1.3 points. These results demonstrate that the expert library itself provides additional adaptation capacity, while adaptive routing is crucial for selecting task-relevant experts under more entangled representations.

\textbf{Prompt-Driven Routing (PDR).} As reported in Table~\ref{tab:ablation}, we further incorporate PDR to dynamically select LoRE experts based on task features extracted from the given prompt. When combined with LoRE, PDR substantially improves MICAS-Cat, reducing Registration CD by 9.1 and increasing Part Segmentation mIoU by 2.8 points. Moreover, it consistently enhances MICAS-Sep across all tasks, including a 1.0 reduction in Denoising CD and a 1.4-point improvement in Part Segmentation mIoU. These results demonstrate that prompt-driven expert selection is critical for achieving effective task specialization.

Overall, LoRE provides substantial performance gains across most tasks, while PDR further enhances robustness and ensures consistent improvements. Without PDR, uniform expert aggregation in LoRE may perturb the entangled input-target point cloud representations in MICAS-Cat, causing negative transfer on Registration. In contrast, PDR adaptively selects task-relevant experts, alleviating such interference and improving task specialization.


\subsection{Analysis Experiment}


\begin{table}[t]
  \centering
  \fontsize{9pt}{11pt}\selectfont
  \setlength{\tabcolsep}{4.4pt}
  \renewcommand{\arraystretch}{0.85}
  \caption{Analysis of the expert library size $K$ in the Low-Rank Expert Library (LoRE).}
  \begin{tabular}{|c||c|c|c|c|c|}
    \hline\thickhline
    \rowcolor{mygray}
    & Library & Rec. & Den. & Reg. & Part Seg. \\
    \rowcolor{mygray}
    \multirow{-2}[-1]{*}{Model} & Size $K$ & CD $\downarrow$ & CD $\downarrow$ & CD $\downarrow$ & mIoU$\uparrow$ \\
    \hline\hline
    \multirow{3}{*}{PromptPath-Sep} & 64 & 3.7 & 4.1 & 3.3 & 88.9 \\
    & 256 & 3.7 & 4.1 & 3.3 & 89.1 \\
    & 1024 & \textbf{3.4} & \textbf{3.3} & \textbf{2.7} & \textbf{89.5} \\
    \hline
  \end{tabular}
  \label{tab:expert_number}
\end{table}

\begin{table}[t]
  \centering
  \fontsize{9pt}{11pt}\selectfont
  \setlength{\tabcolsep}{7.8pt}
  \renewcommand{\arraystretch}{0.85}
   \caption{Robustness analysis of performance changes by replacing retrieved prompts with random prompts.}
  \begin{tabular}{|c||c|c|c|c|}
    \hline\thickhline
    \rowcolor{mygray}
    & Rec. & Den. & Reg. & Part Seg. \\
    \rowcolor{mygray}
    \multirow{-2}[-1]{*}{\makebox[16.3mm][c]{Model}} & CD $\downarrow$ & CD $\downarrow$ & CD $\downarrow$ & mIoU$\uparrow$ \\
    \hline\hline
    \multicolumn{1}{|c||}{MICAS-Sep} & +0.3 & +0.5 & +4.8 & -0.2 \\
    \multicolumn{1}{|c||}{PromptPath-Sep} & \textbf{+0.2} & \textbf{+0.2} & \textbf{+1.3} & -0.2 \\
    \hline
  \end{tabular}
  \label{tab:prompt_mismatch}
\end{table}

\begin{table}[t]
  \centering
  
  \fontsize{9pt}{11pt}\selectfont
  \setlength{\tabcolsep}{3.9pt}
  \renewcommand{\arraystretch}{0.85}
  \caption{Analysis of activated experts in the Low-Rank Expert Library (LoRE). Top-1: the top-1 positive-activation expert; Positive: positive-activation experts; Dual: both positive- and negative-activation experts.}
  \begin{tabular}{|c||c|c|c|c|c|}
  \hline\thickhline
  \rowcolor{mygray}
   & Activated & Rec. & Den. & Reg. & Part Seg. \\
  \rowcolor{mygray}
   \multirow{-2}[-1]{*}{\makebox[17mm][c]{Model}}& Experts & CD $\downarrow$ & CD $\downarrow$ & CD $\downarrow$ & mIoU$\uparrow$ \\
  \hline\hline

  \multirow{3}{*}{\makecell[l]{PromptPath-Cat}} 
  & Top-1  & 4.8 
   & 4.6 
   & 10.0 
  & 87.8 \\
  
  & Positive & \textbf{4.3}
  & \textbf{4.1}
  & 9.7 
  & 89.7 \\

  & Dual & \textbf{4.3}
  & \textbf{4.1}
  & \textbf{7.8}
  & \textbf{89.9}   \\
  
  \hline
  \multirow{3}{*}{\makecell[l]{PromptPath-Sep}}
  & Top-1 & 3.9
  & 4.3
  & 3.2 
  & 86.9  \\

  & Positive & 3.7 
   & 4.1 
  & 3.3 
  & 89.1 \\

  & Dual & \textbf{3.4} 
  & \textbf{3.3} 
  & \textbf{2.7} 
  &  \textbf{89.5} \\
  
  \hline
  \end{tabular}
  \label{tab:analysis_expert}
\end{table}

\begin{table}[t]
  \centering
  
  \fontsize{9pt}{11pt}\selectfont
  \setlength{\tabcolsep}{2.4pt}
  \renewcommand{\arraystretch}{0.85}
  \caption{Comparison Analysis with LoRA-MoE Baselines. Static MoE: fixed-weight expert averaging; Dynamic MoE: prompt-conditioned expert aggregation via a linear router.}
  \begin{tabular}{|c||c|c|c|c|c|}
  \hline\thickhline
  \rowcolor{mygray}
   &  & Rec. & Den. & Reg. & Part Seg. \\
  \rowcolor{mygray}
   \multirow{-2}[-1]{*}{\makebox[17mm][c]{Model}}& \multirow{-2}[-1]{*}{Method} & CD $\downarrow$ & CD $\downarrow$ & CD $\downarrow$ & mIoU$\uparrow$ \\
  \hline\hline
  
  \multirow{3}{*}{\makecell[l]{PromptPath-Cat}} 
  &  Static MoE  & 4.8 
   & 4.7 
   & 13.1 
  & 85.3 \\
  
  &  Dynamic MoE & 4.4
  & 4.2
  & 10.5
  & 89.5  \\

  & Ours & \textbf{4.3}
   & \textbf{4.1}
   & \textbf{7.8}
  & \textbf{89.9} \\

  \hline
  \multirow{3}{*}{\makecell[l]{PromptPath-Sep}} 
  & Static MoE  &  4.2 
   &  4.8 
   &  3.6 
  &  85.5 \\
  
  &  Dynamic MoE & 3.7
  & 3.9
  & 3.1
  & \textbf{89.6}  \\

  & Ours & \textbf{3.4}
  & \textbf{3.3}
  & \textbf{2.7}
  & 89.5 \\

  \hline
  \end{tabular}
  \label{tab:analysis_merge}
\end{table}

\begin{figure}[t]
  \centering
  \includegraphics[width=1.0\linewidth]{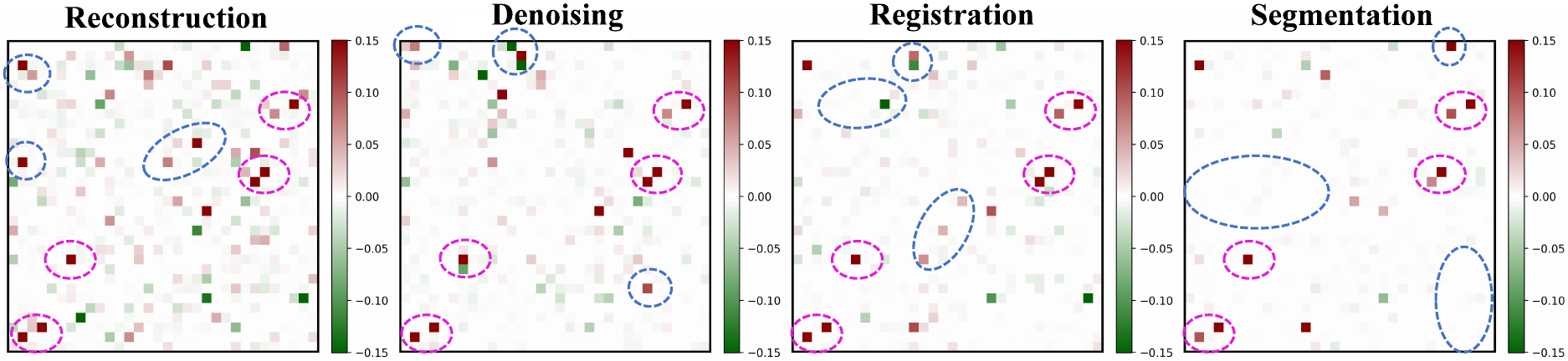}
  \includegraphics[width=1.0\linewidth]{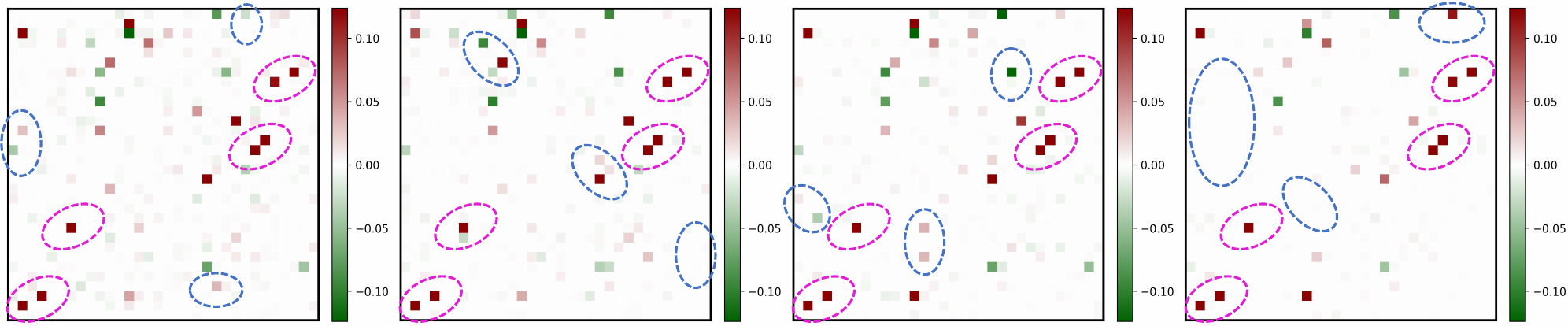}
  \caption{Expert routing heatmap of the low-rank expert library. The top and bottom rows show the routing weights of the up- and down-projection matrices of low-rank experts, respectively. Red and green points represent positively and negatively activated experts, respectively, while magenta and blue ellipses denote task-shared and task-specific experts.}
  \label{analysis_routing1}
\end{figure}

\textbf{1) Sensitivity of Expert Rank Size}. As shown in Table~\ref{tab:analysis_rank}, we evaluate the effect of expert rank by varying the rank from 8 to 32 for PromptPath-Cat and PromptPath-Sep. PromptPath-Cat performs best at rank 32, whereas PromptPath-Sep achieves optimal results at rank 16. These results indicate that an appropriate choice of expert rank is critical for balancing model expressiveness and complexity, with moderate-to-high ranks yielding the most effective performance.

\textbf{2) Sensitivity of Expert Library Size $K$}. As shown in Table~\ref{tab:expert_number}, we vary the number of experts $K$ in the library to investigate the impact of library size. While increasing $K$ from 64 to 256 yields marginal gains, $K=1024$ consistently achieves the best performance across all tasks. This shows that a sufficiently large expert library enables more effective task-specific adaptation through richer expert diversity.

\textbf{3) Robustness to Prompt Selection.} 
To assess the robustness of our method in arbitrary prompts, we replace retrieved prompts with randomly sampled prompts and measure performance changes relative to standard retrieval. As shown in Table~\ref{tab:prompt_mismatch}, random prompt selection degrades both methods, while PromptPath suffers substantially smaller performance drops in Reconstruction, Denoising, and Registration. These results show that our method is robust to prompt selection and can effectively leverage imperfect prompts for task inference.

\textbf{4) Impact of Activated Expert Selection}. We conduct qualitative and quantitative analyses to investigate the impact of activating different experts on model performance. \textbf{First}, Figure~\ref{analysis_routing1} shows that task-specific experts (blue ellipses) receive both positive (red) and negative (green) activations, indicating that they comprise experts with both positive and negative activation. \textbf{Second}, we compare three selection strategies in Table~\ref{tab:analysis_expert}: ``Top-1'', which retains only the expert with the highest positive activation; ``Positive'', which selects all positive-activation experts; and ``Dual'', which selects both positive- and negative-activation experts. Table~\ref{tab:analysis_expert} shows that ``Dual'' outperforms ``Top-1'' and ``Positive'' in PromptPath-Cat and PromptPath-Sep, indicating that discarding negative-activation experts may remove task-relevant modulation signals and result in performance degradation.


\begin{table}[t]
  \centering
  \fontsize{9pt}{11pt}\selectfont
  \setlength{\tabcolsep}{0.9pt}
  \renewcommand{\arraystretch}{0.85}
  \caption{Cosine similarity of prompt routing vectors across tasks. Diagonal entries are within-task similarities. B and A are the up- and down-projection matrices of low-rank experts.}
  \begin{tabular}{|>{\centering\arraybackslash}p{8.0mm}||
    *{3}{>{\centering\arraybackslash}p{8.4mm}}
    >{\centering\arraybackslash}p{8.4mm}||
    *{3}{>{\centering\arraybackslash}p{8.4mm}}
    >{\centering\arraybackslash}p{8.4mm}|}
    \hline\thickhline
    \rowcolor{mygray}
    & \multicolumn{4}{c||}{Up Projection B} & \multicolumn{4}{c|}{Down Projection A} \\
    \rowcolor{mygray}
    \multirow{-2}{8.0mm}{\centering Task} & Rec. & Den. & Reg. & Seg. & Rec. & Den. & Reg. & Seg. \\
    \hline\hline
    Rec. & \textbf{0.44} & -0.09 & -0.12 & -0.19 & \textbf{0.73} & -0.12 & -0.08 & -0.35 \\
    Den. & -0.09 & \textbf{0.42} & -0.18 & -0.12 & -0.12 & \textbf{0.71} & -0.26 & -0.17 \\
    Reg. & -0.12 & -0.18 & \textbf{0.38} & -0.17 & -0.08 & -0.26 & \textbf{0.71} & -0.33 \\
    Seg. & -0.19 & -0.12 & -0.17 & \textbf{0.41} & -0.35 & -0.17 & -0.33 & \textbf{0.56} \\
    \hline
  \end{tabular}
  \label{tab:activation_similarity}
\end{table}

\begin{table}[t]
  \centering
  \fontsize{9pt}{11pt}\selectfont
  \setlength{\tabcolsep}{2.0pt}
  \renewcommand{\arraystretch}{0.90}
  \caption{Analysis of cross-task generalization using Part Segmentation as the evaluation task.}
  \begin{tabular}{|>{\centering\arraybackslash}m{23mm}||>{\centering\arraybackslash}p{26mm}|>{\centering\arraybackslash}p{28.5mm}|}
    \hline\thickhline
    \rowcolor{mygray}
    \raisebox{-1.50ex}{Model} & Training All Tasks w/ Part Seg. mIoU$\uparrow$ & Training All Tasks w/o Part Seg. mIoU$\uparrow$ \\
    \hline\hline
    MICAS-Sep & 86.8 & 77.5 \\
    PromptPath-Sep & \textbf{89.5} & \textbf{85.8} \\
    \hline
  \end{tabular}
  \label{tab:heldout_task}
\end{table}

\textbf{5) Comparison with LoRA-MoE Baselines}. As shown in Table~\ref{tab:analysis_merge}, we compare two existing LoRA-MoE variants with the same number of LoRA experts: static MoE and dynamic MoE. Static MoE removes the routing mechanism and uniformly aggregates all experts with equal weights of $1/K$, whereas dynamic MoE combines activated experts using weights generated by a standard linear router conditioned on the prompt. We find that static MoE achieves the lowest performance, while dynamic MoE improves upon it but generally remains inferior to our method. These results demonstrate that the performance gains arise from prompt-driven expert composition rather than simply increasing LoRA capacity, and further validate the effectiveness of our proposed prompt-driven routing (PDR) over standard linear routing.

\textbf{6) Interpretability of Dynamic Routing.} We conduct qualitative and quantitative analyses to investigate the interpretability of dynamic routing. \textbf{First}, we visualize the activation weights of low-rank experts in the attention layer of decoder layer 0 for PromptPath-Cat on a specific inference instance. As shown in Figure~\ref{analysis_routing1}, the heatmaps of the up- and down-projection matrices reveal both shared and task-specific expert activations generated by the prompt-driven routing (PDR) module. The purple circles highlight activation regions shared across tasks, while the blue circles denote task-specific regions, demonstrating that PDR adaptively modulates expert contributions according to task prompts and forms task-aware computation pathways. \textbf{Second}, we quantify the interpretability using cosine similarity between routing vectors across tasks. As shown in Table~\ref{tab:activation_similarity}, routing vectors exhibit higher within-task and lower cross-task similarities for both projection matrices, indicating that prompts from the same task induce consistent routing signals, whereas prompts from different tasks generate distinct routing signals. Together, these results demonstrate that PDR learns task-discriminative routing signals and enables interpretable computation pathway formation.

\textbf{7) Capability of Model Generalization.} We evaluate the generalization capability of PromptPath across diverse domains and tasks. \textbf{(i) Cross-Domain Generalization.} Following Condenser~\cite{Condenser}, we train on COCO-5i~\cite{lin2014microsoft} and evaluate on Pascal-5i~\cite{shaban2017one}. As shown in Table~\ref{tab:sota_ni}, PromptPath achieves superior transfer performance, outperforming Condenser by 0.89 mIoU. \textbf{(ii) Cross-Task Generalization.} We train on Reconstruction, Denoising, and Registration while holding out Part Segmentation as an unseen task for evaluation. PromptPath-Sep achieves 85.8 mIoU on the held-out task, surpassing MICAS-Sep by 8.3 points (Table~\ref{tab:heldout_task}). Moreover, it incurs only a 3.7 mIoU drop compared with the all-task setting, whereas MICAS-Sep suffers a larger degradation of 9.3 points. These results demonstrate the superior ability of PromptPath to generalize across unseen domains and task variations.

\textbf{8) Computational and Parameter Efficiency.}
\label{sec:supp_efficiency}

\begin{table}[t]
  \centering
  \fontsize{9pt}{11pt}\selectfont
  \setlength{\tabcolsep}{4.7pt}
  \renewcommand{\arraystretch}{1.20}
  \caption{Inference overheads are measured with a batch size of 16 on a single V100 GPU. We use MICAS-Sep as the baseline model and conduct a comprehensive comparison with our PromptPath-Sep model across multiple metrics.}
  \begin{tabular}{|c||c|c|c|c|}
    \hline\thickhline
    \rowcolor{mygray}
    Model & \makecell{Latency\\(ms/sample)} & \makecell{GPU Mem.\\(MB)} & FLOPs & \makecell{Wall-Clock\\Time (s)} \\
    \hline\hline
    \multicolumn{1}{|c||}{Baseline} & 44.6 & 3512 & 23.27G & 3986.9 \\
    \hline
    \multicolumn{1}{|c||}{Ours} & 45.9 & 4384 & 25.09G & 4093.3 \\
    \hline
  \end{tabular}
  \label{tab:analysis_cost}
\end{table}

\begin{table}[t]
  \centering
  \fontsize{9pt}{11pt}\selectfont
  \setlength{\tabcolsep}{2.1pt}
  \renewcommand{\arraystretch}{1.05}
  \caption{Trainable-parameter comparison with a Mixture-of-LoRA (MoLA) design~\cite{feng2024mixture}.}
  \begin{tabular}{|c||>{\centering\arraybackslash}p{19mm}|*{3}{>{\centering\arraybackslash}p{13.5mm}|}}
    \hline\thickhline
    \rowcolor{mygray}
    \multicolumn{1}{|c||}{Model} & \makecell{Expert number\\($\times$ blocks)} & \makecell{One\\expert} & \makecell{All\\experts} & \makecell{Routing\\module} \\
    \hline\hline
    MoLA & 8 & 143M & 1144M & -- \\
    PromptPath & \makecell{64/256/1024\\($\times 4$)} & 0.0375M & \makecell{9/38/\\151M} & 73M \\
    \hline
  \end{tabular}
  \label{tab:param_comparison}
\end{table}

Tables~\ref{tab:analysis_cost} and \ref{tab:param_comparison} provide detailed computational and parameter efficiency analyses. PromptPath-Sep yields gains over MICAS-Sep with limited inference overhead: latency rises from $44.6$ to $45.9$ ms/sample and wall-clock time from $3986.9$ to $4093.3$ s. Although prompt-aware routing and fusion increase FLOPs and memory use, the runtime impact remains small. Each expert has 0.0375M trainable parameters; the 1024-expert library across four decoder blocks and PDR contain 151M and 73M parameters, respectively. Thus, PromptPath assembles task-specific capacity from experts while remaining practical for inference.


\section{Conclusion}


We identify a \textbf{shallow task adaptation issue} in ICL methods, where prompts support implicit task inference through input conditioning. We propose \textbf{PromptPath}, a prompt-adaptive ICL framework that uses prompt-driven routing to activate and compose low-rank experts, enabling task-specific, interpretable computation. Across 2D and 3D benchmarks, PromptPath outperforms state-of-the-art ICL methods, demonstrating the value of computation-level prompt conditioning for task adaptation and generalization.

\section{Limitations}

PromptPath performs well on evaluated 2D and 3D in-context learning benchmarks, but robustness to real-world shifts---especially severe out-of-distribution changes in data sources, noise, corruptions, and tasks---remains unverified. Rank analysis is capped at 32 by computation; larger ranks or expert libraries may improve performance but require efficiency--performance evaluation.

{
    \small
    \bibliographystyle{ieeenat_fullname}
    \bibliography{main}
}

\end{document}